\documentclass[a4paper]{svproc}
\usepackage{style}
\usepackage{url}
\usepackage{wrapfig}
\usepackage{soul}
\usepackage[absolute,overlay]{textpos}

\begin{document}
\mainmatter              
\title{Design and Realization of a \\Benchmarking Testbed for Evaluating\\Autonomous Platooning Algorithms}
\titlerunning{Platoon Testbed}  
%
\author{Michael~H.~Shaham \and Risha Ranjan \and Engin K{\i}rda \and Ta\c{s}k{\i}n Pad{\i}r}
\authorrunning{Michael~H.~Shaham et al.} 
\institute{Northeastern University, Boston, MA 02115, USA,\\
\email{shaham.m@northeastern.edu},\\ 
\texttt{https://robot.neu.edu/}
}
\maketitle              

\begin{textblock*}{120mm}(45mm, 5mm)
\begin{center}
\definecolor{darkgray}{rgb}{.4, .4, .4}
\textcolor{darkgray}{
Accepted for publication at ISER 2023. Please cite as follows:\\
M. H. Shaham, R. Ranjan, E. K{\i}rda, T. Pad{\i}r, ``Design and Realization of a Benchmarking Testbed for Evaluating Autonomous Platooning Algorithms,'' in International Symposium on Experimental Robotics, 2023.
}
\end{center}
\end{textblock*}

\begin{abstract}  
Autonomous vehicle platoons present near- and long-term opportunities to enhance operational efficiencies and save lives. The past 30 years have seen rapid development in the autonomous driving space, enabling new technologies that will alleviate the strain placed on human drivers and reduce vehicle emissions. This paper introduces a testbed for evaluating and benchmarking platooning algorithms on 1/10th scale vehicles with onboard sensors. To demonstrate the testbed's utility, we evaluate three algorithms, linear feedback and two variations of distributed model predictive control, and compare their results on a typical platooning scenario where the lead vehicle tracks a reference trajectory that changes speed multiple times. We validate our algorithms in simulation to analyze the performance as the platoon size increases, and find that the distributed model predictive control algorithms outperform linear feedback on hardware and in simulation.
\end{abstract}

\section{Introduction}
\label{sec:intro}

\subsection{Motivation}
\label{subsec:motivation}

Autonomous driving technologies have the potential to improve safety, reduce fuel emissions \cite{SARTRE2011}, reduce stress, and more \cite{rojas2020autonomous}. In this work, we focus on autonomous vehicle platoons, or groups of autonomous vehicles driving together, which also have the potential to increase highway capacities and reduce travel time. Since the 1990s, there have been numerous efforts to demonstrate platooning technologies including the California PATH program demonstration \cite{PathDemo97}, the 2011 Grand Cooperative Driving Challenge \cite{GCDC2011}, the Energy ITS project \cite{EnergyITS2011}, and the SARTRE platooning program \cite{SARTRE2011}. Though these demonstrations highlight the feasibility of platoons for the trucking industry or automated highway systems, they often evaluate only a single algorithm, and there is still a need to benchmark algorithms to compare safety and performance \cite{Padir19}.

Validation platforms for autonomous driving algorithms on hardware have been introduced, but these systems typically focus on autonomous racing \cite{MITRacecar2017,AutoRally2016,F1Tenth2020} or coordinated planning and networked control \cite{CambridgeMinicar2019,CPMLab2021}. With the exception of \cite{lee2022cyclops} which focuses on heavy-duty vehicles, there remains a lack of focus on vehicle platooning. This work introduces a research platform that builds on the widely used open source F1Tenth system \cite{F1Tenth2020} to test autonomous platooning algorithms. Using our platform, we evaluate three different algorithms and perform further analysis of these algorithms in simulation.

\subsection{Related Work}
\label{subsec:related_work}

The platooning problem was studied as early as the 1960s \cite{StringVehicles1966}, but it was not until the California PATH demonstration and associated research in the 1990s when platooning began to receive considerable attention \cite{PathDemo97}. Since then, many theoretical developments have provided insight into the performance of vehicle platooning algorithms. These insights typically relate to how the information flow topology, distributed controller, and spacing policy contribute to the overall behavior \cite{FourComponent2015}. 

For analysis of platoons, there are two common performance metrics used: string stability \cite{StringStab1996,feng_string_2019} and (asymptotic) stability of the platoon. String stability ensures disturbances, typically due to the head vehicle's motion, will not propagate down the string of vehicles in the platoon. An algorithm that is stable follows the usual control theoretic definition: subject to some disturbance, the vehicles in the platoon will eventually be able to reach zero error. Platoons typically use two different inter-vehicle spacing policies: (1) constant distance headway (CDH) where each vehicle maintains a constant distance to its predecessor and (2) constant time headway (CTH) where each vehicle maintains a distance based on the time required to reach its predecessor.

Linear feedback controllers have been widely used to obtain insightful theoretical results. In \cite{ZhengStabilityScalability2016}, the authors analyze the effect of the information flow topology on the stability of a platoon using a linear feedback controller and a CDH policy, and show how performance degrades significantly as platoon size increases. The type of controller along with the spacing policy can also significantly impact the performance of a platoon. In \cite{DisturbanceProp2004}, authors prove that the combination of a linear feedback policy, a predecessor follower topology (each vehicle has access to information only from the preceding vehicle), and a CDH policy leads to string instability. However, under the same conditions but with a CTH policy, string stability is guaranteed \cite{CACCDesign2010}. Similarly, in \cite{alam_suboptimal_2011}, the decentralized structure of the platooning problem and linear quadratic regulator theory are exploited to design a string stable linear feedback controller using a CTH policy.

By assuming vehicles can communicate with one another, it is possible to design distributed model predictive control (DMPC) algorithms. In \cite{DistRHCPlatoon2012}, the authors present the first DMPC algorithm that guarantees both asymptotic stability and string stability. Building on this, the authors in \cite{DistMPCHetero2017} guarantee stability (but not string stability) of the platoon without following vehicles knowing a priori the desired set point (velocity) of the platoon. Both \cite{DistRHCPlatoon2012} and \cite{DistMPCHetero2017} assume vehicles share their planned trajectory with neighboring vehicles, which is used at the next timestep. More recently, \cite{qiang_distributed_2023} develop a DMPC method that allows vehicles to share their true optimal planned trajectory multiple times at each timestep, while also using collision constraints to guarantee safety.

\subsection{Problem Statement}
\label{subsec:problem_statement}

A platoon consists of a string of vehicles following one another. The goal of the platoon is to have each vehicle drive at the same speed as the leader while maintaining a desired spacing relative to its neighbors (we focus on a CDH policy). Most platooning research in the literature contribute theoretical results and a simulation experiment to validate the theory, without performing any validation on hardware (and often without comparing performance to other algorithms). To evaluate platooning controllers, a benchmarking testbed needs to be developed to compare results and analyze advances and trade-offs. The testbed should use vehicles equipped with sensors that provide inter-vehicle sensing and communication. The vehicles must also be capable of following trajectories typical of highway-driving scenarios and drive at a sufficiently large range of speeds to track these trajectories.

\section{Technical Approach}
\label{sec:technical_approach}

\subsection{Hardware}
\label{subsec:hardware}

\begin{wrapfigure}{r}{0.39\textwidth}
    \vspace{-8.0mm}
    \includegraphics[width=\linewidth]{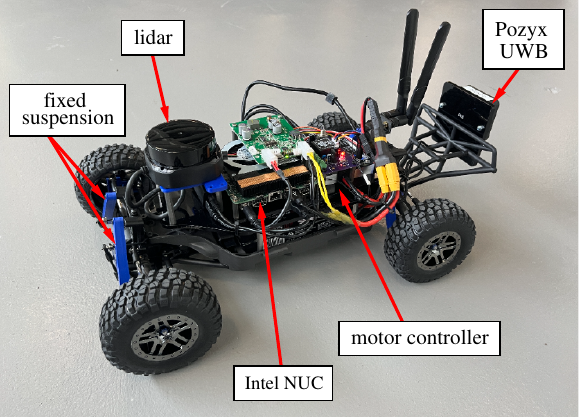}
    \vspace{-6.0mm}
    \caption{One of the vehicles used in our platoon, which is a modified version of the F1Tenth vehicle.}
    \label{fig:convoy_vehicle}
    \vspace{-6.0mm}
\end{wrapfigure}

The vehicles (Figure \ref{fig:convoy_vehicle}) used in our experiments are modified versions of the standard F1Tenth vehicle \cite{F1Tenth2020}. We introduce the following modifications: using an Intel NUC instead of an NVIDIA Jetson Xavier NX for faster compute, adding an AS5047P magnetic motor position sensor for better low-speed performance, using fixed suspension rods to reduce jitter due to the suspension, and adding a Pozyx ultra-wideband (UWB) radio frequency device to obtain inter-vehicle distance measurements. The Pozyx UWB device is reported to be accurate up to $\pm10$~cm. In our experiments, we have found its error to be roughly Gaussian with a standard deviation of 4.5~cm. All perception, planning, and control computations are completely decentralized and are performed on the Intel NUC using ROS 2 \cite{ROS22022}. The testing environment used for hardware experiments is a 10.2~m$\times$6.6~m reconfigurable, oval race track.

Since this paper focuses on multi-vehicle longitudinal control, we briefly summarize the local planning each vehicle performs for lateral control. Each vehicle is equipped with a 2D lidar---either an RPLiDAR S2 (follower vehicles) or a Hokuyo UST-10LX (leader vehicle). The lidar provides data in the form of a pointcloud, which we segment using DBSCAN \cite{DBSCAN1996} to find the sets of points corresponding to the left and right walls. The points associated with the left and right walls are then used to estimate a polynomial centerline (in coordinates relative to the local vehicle's frame), which is passed as input to the pure pursuit algorithm \cite{PurePursuit1992}. Note that all planning is performed locally, \ie, the vehicles have no notion of global position within the course; they only know their relative position to their predecessor in the platoon. 

\subsection{Control Algorithms}
\label{subsec:control_algorithms}

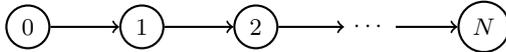
\begin{figure}
    \vspace{-4.0mm}
    \centering
    \begin{tikzpicture}[node distance={15mm}, thick, main/.style = {draw, circle}]
        \node[main] (0) {$0$};
        \node[main] (1) [right of=0] {$1$};
        \draw[->] (0) -- (1);
        \node[main] (2) [right of=1] {$2$};
        \draw[->] (1) -- (2);
        \node (k) [right of=2] {$\cdots$};
        \draw[->] (2) -- (k);
        \node[main] (N) [right of=k] {$N$};
        \draw[->] (k) -- (N);
    \end{tikzpicture}
    \vspace{-2.0mm}
    \caption{Predecessor follower topology. The directed edge $i \to j$ indicates vehicle $j$ has access to (or receives information from) vehicle $i$. Vehicles are indexed by $0, 1, \ldots, N$.}
    \label{fig:pf_topology}
    \vspace{-9.0mm}
\end{figure}

The platoon consists of $N+1$ vehicles, indexed by $0, 1, \ldots, N$, where vehicle 0 is the leader. We analyze and compare a linear feedback and two DMPC controllers using a predecessor follower topology, depicted in \cref{fig:pf_topology}. We use a constant spacing policy, and the goal of each following vehicle is to match the velocity of and maintain a constant distance to its predecessor, \ie, 
\begin{equation}
\label{eq:platoon_asym_stab}
    \begin{array}{rl}
        p_i(t) & \to p_{i-1}(t) - d_\text{des} \\
        v_i(t) & \to v_{i-1}(t) \\
    \end{array}
    \qquad \text{as} \ t \to \infty, \text{ for all } i = 1, \ldots, N
\end{equation}
where $p$ and $v$ are position and velocity, respectively, and $d_\text{des}$ is the desired inter-vehicle distance. \Cref{eq:platoon_asym_stab} guarantees asymptotic stability---we do not focus on string stability in this work. Note that if \cref{eq:platoon_asym_stab} is achieved, each vehicle also matches the desired speed of and the desired spacing relative to the leader.

\subsubsection{Vehicle dynamics.}

We model each vehicle's dynamics as
\begin{equation}
\label{eq:hw_dyn}
\begin{gathered}
    x_i(k+1) = A_i x_i(k) + B_i u_i(k) \\
    A_i = \bmat{1 & \dt \\ 0 & 1 - \frac{\dt}{\tau_i}}
    \quad B_i = \bmat{0 \\ \frac{\dt}{\tau_i}}
\end{gathered}
\qquad i = 0, \ldots, N
\end{equation}
where $k$ is the discrete sampling instance, $\dt$ is the discretization time and $\tau_i$ is the inertial delay of the longitudinal dynamics. The state $x_i = (p_i, v_i)$ is the position and velocity and the control input $u_i$ is the desired velocity. It is more common to use a dynamics model where the state $x_i$ includes the vehicle's acceleration and the input $u_i$ is the desired acceleration. However, we have found the F1Tenth vehicles perform better when specifying a desired velocity (corresponding to motor revolutions per minute) instead of an acceleration (corresponding to motor current).

\subsubsection{Linear feedback algorithm.}

In the linear feedback case, each vehicle measures or receives the distance to and velocity of its predecessor. The distance measurements are obtained via the UWB device and the speed measurements are shared using ROS 2 and Wi-Fi. Real vehicles are commonly equipped with radar or a similar sensor to estimate the distance and speed of neighboring vehicles. The desired velocity input for the linear feedback controller is selected using 
\begin{equation*}
    \label{eq:lfbk}
    u_i(t) = k_p (p_{i-1}(t) - p_i(t) + d_\text{des}) + k_v (v_{i-1}(t) - v_i(t)), \quad i = 1, \ldots, N
\end{equation*}
where $k_p, k_v > 0$ are predefined gains. See \cite{ZhengStabilityScalability2016} for further details and stability analysis of the continuous-time version of this controller when using the typical acceleration-input version of the dynamics. The results in \cite{ZhengStabilityScalability2016} can be extended to systems using the dynamics given by \cref{eq:hw_dyn}. 

\subsubsection{Distributed model predictive control algorithms.}

For the DMPC controllers, at each timestep $t$, each vehicle receives the assumed trajectory of its predecessor, which is calculated at the previous timestep $t-1$. Let $x_i^a(0), \ldots,$ $x_i^a(H)$ denote the assumed trajectory over the planning horizon $H$ for vehicle $i$ at timestep $t$. Then at each timestep $t$, the DMPC controller solves
\begin{equation}
    \label{eq:dmpc}
    \def\arraystretch{1.3}
    \arraycolsep=2.5pt
    \begin{array}{ll}
        \underset{u_i^p, x_i^p}{\mbox{minimize}} & \sum_{k=0}^{H-1} l_i\left(x_i^p(k), u_i^p(k), x_i^a(k), x_{i-1}^a(k)\right) \\
        \mbox{subject to} & x_i^p(0) = x_i(t) \\
        & x_i^p(k+1) = A_i x_i^p(k) + B_i u_i^p(k), \quad k = 0, \ldots, H-1 \\
        & \abs{v_i^p(k+1) - v_i^p(k)} \leq \dt \cdot a_{i, \text{max}}, \quad k = 0, \ldots, H-1 \\
        & v_{i, \text{min}} \leq v_i^p(k) \leq v_{i, \text{max}}, \quad k = H, \ldots, H \\
        & x_i^p(H) = x_{i-1}^a(H) - \tilde d_{\text{des}} \\
        & u_i^p(H-1) = v_{i-1}^a(H)
    \end{array}
\end{equation}
where $\tilde d_\text{des} = (\tilde d, 0) \in \reals^2$. This is the same formulation proposed in \cite{DistMPCHetero2017} but with slightly altered constraints to account for the dynamics model we use.

We analyze two different options for the cost function $l_i$. The first uses the squared weighted 2-norm to penalize errors:
\begin{equation}
\label{eq:squared_2_norm_cost}
    l_i(k) = \normm{x_i^p(k) - x_i^a(k)}_{F_i}^2 + \normm{x_i^p(k) - x_{i-1}^a(k) + \tilde d_{des}}_{G_i}^2 + \normm{u_i^p(k) - \tilde u}_{R_i}^2
\end{equation}
where $F_i, G_i, R_i \succ 0$ and $\norm{z}_Q^2 = z^\tp Q z$. The second option uses the 1-norm:
\begin{equation}
\label{eq:1_norm_cost}
    l_i(k) = s_i \normm{x_i^p(k) - x_i^a(k)}_1 + q_i \normm{x_i^p(k) - x_{i-1}^a(k) + \tilde d_{des}}_1 + r_i \normm{u_i^p(k) - \tilde u}_1
\end{equation}
where $s_i, q_i, r_i > 0$ are scalar variables weighting each objective. The first term in these cost functions, which we refer to as the \textit{move suppression term}, penalizes the deviation between vehicle $i$'s predicted and assumed trajectories. The second term, which we refer to as \textit{predecessor relative error term}, penalizes the error between vehicle $i$'s predicted trajectory and its predecessor's assumed trajectory. We will refer to the DMPC algorithm (\ref{eq:dmpc}) using \cref{eq:squared_2_norm_cost} as \textit{squared 2-norm} (or $\norm{\cdot}_2^2$) DMPC and the algorithm using \cref{eq:1_norm_cost} as \textit{1-norm} (or $\norm{\cdot}_1$) DMPC. Note that the timestep-shifted version of the optimal trajectory computed at timestep $t$ is used as the assumed trajectory for timestep $t+1$. See \cite{DistMPCHetero2017} for further details about the implementation of the algorithm.

The inequality constraints of \cref{eq:dmpc} enforce velocity and acceleration limits. The equality constraints of \cref{eq:dmpc} serve two purposes: (1) ensure dynamic feasibility of the trajectory and (2) enforce terminal constraints to ensure recursive feasibility of the DMPC controller. If we denote the solution of this optimization problem as $u_i^\star$, then the DMPC controller implements $u_i^\star(0)$ and repeats this process at the next timestep after receiving the assumed trajectory of its predecessor. Since our controller uses the desired velocity as the input, we use $\tilde u = v_{i}(t)$, so each vehicle penalizes velocity input that deviate from its current velocity.

The DMPC algorithm we use is based on the formulation used in \cite{DistMPCHetero2017}, which uses the cost function 
\begin{equation}
\label{eq:2_norm_cost}
    l_i(k) = \normm{x_i^p(k) - x_i^a(k)}_{F_i} + \normm{x_i^p(k) - x_{i-1}^a(k) + \tilde d_{des}}_{G_i} + \normm{u_i^p(k) - \tilde u}_{R_i}.
\end{equation}
We will refer to this as 2-norm DMPC. The theoretical basis for using the weighted 2-norm, which is provided in \cite{DistMPCHetero2017}, is given in the following theorem, which is a slight adaptation of theorem 5 from \cite{DistMPCHetero2017}.
\begin{theorem}
\label{thm:suff_cond}
    Suppose a platoon uses a predecessor follower topology, and each vehicle has dynamics given by \cref{eq:hw_dyn}. If each vehicle uses the DMPC controller that requires solving \cref{eq:dmpc} with the cost function given by \cref{eq:2_norm_cost}, then the platoon's dynamics are asymptotically stable (\ie, \cref{eq:platoon_asym_stab} is achieved) if 
    \begin{equation*}
        F_i \succeq G_{i+1}, \quad i = 0, 1, \ldots, N-1.
    \end{equation*}
\end{theorem}

Recall that $F_i$ is related to the move suppression term of the $i$th vehicle, and $G_{i+1}$ is the predecessor relative error term of the $i$th vehicles predecessor. Intuitively, this result says that each vehicle must place at least as much weight on maintaining its assumed trajectory as its predecessor does. \Cref{thm:suff_cond} was derived for a nonlinear vehicle dynamics model that is equivalent to a linear dynamics model (where the state includes the acceleration) after using the feedback linearization technique. However, it is not too difficult to adjust the proof to work for the dynamics given by \cref{eq:hw_dyn} with the constraints in \cref{eq:dmpc}.

Though not formally analyzed in \cite{DistMPCHetero2017}, it is relatively easy to extend \cref{thm:suff_cond} to the case of the 1-norm cost function given by \cref{eq:1_norm_cost}. The proof in \cite{DistMPCHetero2017} requires only the use of norm inequalities, and thus also applies to the case of the 1-norm, but with minor adjustments to handle the scalar terms instead of the weight matrices. With that said, we extend \cref{thm:suff_cond} to the case of the 1-norm DMPC with the following sufficient condition to guarantee platoon asymptotic stability:
\begin{equation}
\label{eq:l1_suff_cond}
    s_i \geq q_{i+1}, \quad i = 0, 1, \ldots, N-1.
\end{equation}

So far, we have not been able to determine any sufficient conditions that prove the DMPC algorithm with the squared 2-norm cost function (\ref{eq:squared_2_norm_cost}) is stable. Our results, however, will show that the algorithm performs well in practice, so we believe it is worthwhile to consider even if it is not (yet) provably stable. Further, we decide to use the cost functions given by \cref{eq:squared_2_norm_cost} and \cref{eq:1_norm_cost} as opposed to the \cref{eq:2_norm_cost} because these cost functions lead to optimization problems that can be formulated as a quadratic program (QP) or a linear program (LP), respectively. These formulations allows us to plan over a much larger time horizon ($H = 100$ in our experiments) and alleviate feasibility issues due to the terminal equality constraints. For the weighted 2-norm cost function (\ref{eq:2_norm_cost}), we were unable to scale the time horizon above roughly $H = 20$, and ran into feasibility issues when testing on hardware where the controller needs to run at a rate of at least 10\,Hz. To solve the optimization problems, we used CVXPY \cite{diamond2016cvxpy} with OSQP \cite{stellato2020osqp} (for the QP) or with ECOS \cite{domahidi2013ecos} (for the LP). With these tools, we were able to solve the optimization problem (\ref{eq:dmpc}) at rates faster than 100\,Hz.

\section{Experimental Results and Insights}
\label{sec:experiments}

\subsection{Hardware results}
\label{subsec:hw_results}

\begin{figure}
    \vspace{-6.0mm}
    \centering
    \includegraphics{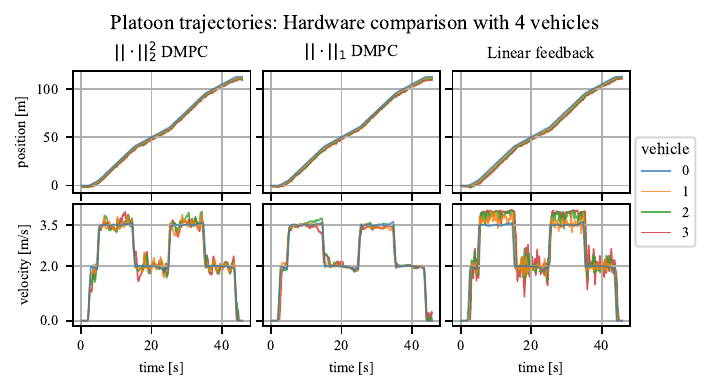}
    \vspace{-5.0mm}
    \caption{Results depicting the four-vehicle platoon trajectories when using three different algorithms on the hardware described in \cref{subsec:hardware}. The columns show the performance of the squared 2-norm DMPC, 1-norm DMPC, and the linear feedback algorithms. The position of the lead vehicle was calculated by integrating its velocity, and the positions of the following vehicles was calculated using the preceding vehicle's position offset by the distance measurement from the Pozyx UWB.}
    \label{fig:exp_comp}
    \vspace{-3.0mm}
\end{figure}

To compare the performance of the two DMPC controllers and the linear feedback controller, we conducted an experiment where the lead vehicle starts at 0\,m/s velocity, accelerates to 2\,m/s, twice alternates accelerating to 3.5\,m/s and back down to 2\,m/s, before finally coming to a stop. The experiment was designed to test how well the platoon handles sudden changes in speed (\eg, as in highway driving due to speed limit changes) and its ability to follow the leader from start to stop. \Cref{fig:exp_comp} shows results from this experiment with four vehicles (one leader and three followers) for each of the three controllers. We used a longitudinal time delay of $\tau_i = 0.3$ for each vehicle (determined experimentally using a least-squares regression on experimental data) and a desired spacing of 1\,m. For each of the DMPC algorithms, we used the identity matrix (weighting matrices in \cref{eq:squared_2_norm_cost}) or the scalar $1$ (scalar weights in \cref{eq:1_norm_cost}) for all controller parameters, which satisfies conditions given in \cref{thm:suff_cond} and \cref{eq:l1_suff_cond}. For linear feedback, we used $k_p = 1$ and $k_v = 2$, as we found these values gave the smoothest vehicle trajectories during testing.

\begin{figure}
    \vspace{-5.0mm}
    \centering
    \includegraphics{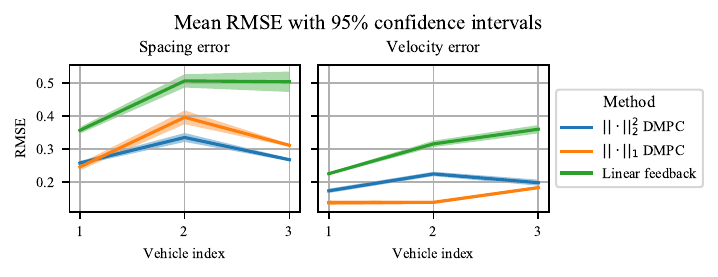}
    \vspace{-8.0mm}
    \caption{A comparison of results over ten trials of each algorithm.}
    \label{fig:rms_comp}
    \vspace{-3.0mm}
\end{figure}

To quantitatively compare the performance between each of the three controllers, we repeated the experiment shown in \cref{fig:exp_comp} 10 times for each controller, completing a total of 30 experiments. We use the root mean square error (RMSE) of the platoon's spacing and velocity trajectories to compare the three controllers. The spacing and velocity errors are given by $p_i - p_{i-1} + d_\text{des}$ and $v_i - v_{i-1}$, respectively. For each experiment, we calculate the mean spacing and velocity RMSE for each following vehicle. We then calculated the mean and standard deviation of these mean RMSE values over the ten experiments, and calculated a 95\% confidence interval for the mean RMSE of each vehicle's position and velocity. The results are shown in \cref{fig:rms_comp}. As we can see, the two DMPC controllers clearly outperform the linear feedback controller.

In terms of spacing error, the squared 2-norm DMPC performs better than the 1-norm DMPC for the two furthest followers, and the two algorithms have comparable performance for the first follower. The 1-norm DMPC controller performed the best in terms of velocity error. One thing our analysis on hardware is unable to tell us is how well these algorithms scale as the size of the platoon increases. To investigate this issue, we conducted experiments in simulation, discussed below.

\subsection{Simulation results}

\begin{figure}
    \vspace{-3.0mm}
    \centering
    \includegraphics{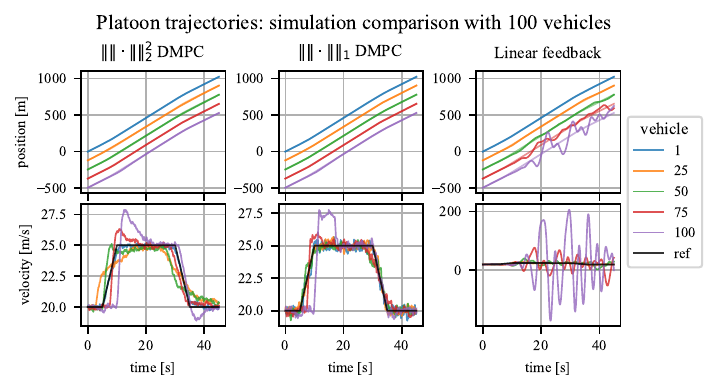}
    \vspace{-8.0mm}
    \caption{Results from one simulated experiment with a 100-vehicle platoon. The reference velocity trajectory the leader follows is shown as the black curve in the velocity plots (bottom row). In the position plots, the vehicle 1 trajectory is indistinguishable from the reference. Note the difference in $y$-axis scales for the bottom row of plots. We also plot the desired positions of each vehicle relative to the leader as faded curves in the top row, and this is most visible in the linear feedback case.}
    \label{fig:sim_scaled_traj}
    \vspace{-6.0mm}
\end{figure}

To analyze how well these algorithms perform as the size of the platoon increases, we simulate a platooning experiment with $N = 100$ following vehicles. In the experiment, the leader follows a reference trajectory that starts at 20\,m/s, accelerates to 25\,m/s and waits for 30 seconds, then decelerates back down to 20\,m/s. Trajectory results for a few vehicles are shown in \cref{fig:sim_scaled_traj}. We use $\tau_i = 0.3$ for each vehicle and set the desired spacing at 5 meters. We also used the same controller parameters that were used in the hardware experiments for the simulation study. For all simulation experiments, we add zero-mean Gaussian noise with a covariance of $.3I$ to the dynamics and with a standard deviation of 4.5\,cm to the inter-vehicle spacing sensing (to match the error encountered by the Pozyx UWB device).

\Cref{fig:sim_scaled_traj} shows that the two DMPC algorithms are able to remain close to the desired platoon trajectory without requiring unsafe trajectories. The 100th vehicle in the platoon for each of these controllers reaches a peak velocity of almost 28\,m/s before settling to the desired velocity of 25\,m/s during the acceleration portion of the experiment. They also perform well when decelerating. Despite this, it is still clear that the performance relative to the true desired goal of maintaining the leader's velocity profile degrades as we move further down the platoon (since platoons further from the leader need to achieve higher velocities to maintain performance). Though it is not shown in these figures, when using the DMPC controllers, each of the vehicles maintained a spacing error of less than roughly one meter, meaning the vehicles never came close to crashing when using a desired spacing of 5 meters. It is not clear to what extent we would need to scale the platoon size $N$ before seeing a collision between two vehicles under this problem setup.

The performance of the linear feedback algorithm, however, quickly degrades as the platoon size increases. Though we saw that performance on hardware was reasonable, we can see from the position trajectories that if we continue to allow the platoon size to increase, the vehicles will crash if the desired spacing is not large enough. Note that we allowed the linear feedback vehicles to apply any unconstrained input and set no bounds on the state. Thus, the 100th vehicle in the platoon was allowed to command large control inputs in an attempt to stabilize. Distributed linear feedback controllers using a CTH policy would alleviate the issues we see here, but we do not investigate this problem setting in this work.

\begin{figure}
    \vspace{-2.0mm}
    \centering
    \includegraphics{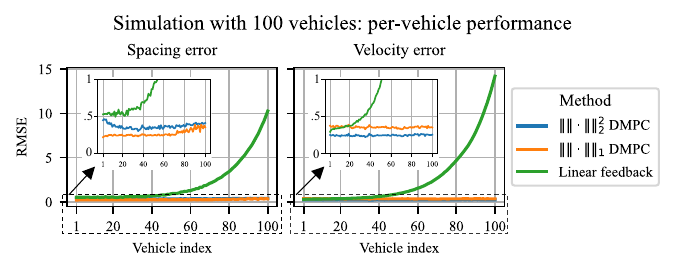}
    \vspace{-5.0mm}
    \caption{Results showing the mean RMSE calculated after repeating the experiment shown in \cref{fig:sim_scaled_traj} ten times. The zoomed in portion of each plot shows the DMPC results more clearly and the platoon size at which linear feedback begins to diverge.}
    \label{fig:sim_rms}
    \vspace{-4.0mm}
\end{figure}

Similar to the hardware experiments, we repeat the simulation experiment ten times and investigate the average performance of each vehicle in the platoon with respect to RMSE. \Cref{fig:sim_rms} shows the results of this analysis. In line with the results in \cite{ZhengStabilityScalability2016}, the performance of linear feedback quickly degrades as the size of the platoon increases. In fact, performance degrades quadratically as the platoon size $N$ increases, and would not be practical above roughly $N = 25$. Optimal control methods could be used to select better gain values, but for the sake of comparison we used the same values in simulation and on hardware.

Unlike the linear feedback controller, the DMPC controller performance does not degrade significantly as the platoon size increases up to $N = 100$ following vehicles. This is evident in the zoomed-in plots of \cref{fig:sim_rms}, where we can see the RMSE does not get worse as we move to the end of the platoon. In the spacing RMSE, however, we can see that the performance appears to be trending worse as we move further down the platoon, but this trend may not become dramatic enough until the platoon becomes much larger, potentially beyond practical platoon sizes. Finally, it is interesting to note that the better performance of the 1-norm compared to the squared 2-norm DMPC controller on hardware did not translate to simulation.

\section{Conclusion}
\label{sec:conclusion}

In this work, we introduced a benchmarking testbed to evaluate autonomous platooning algorithms. We introduced two DMPC algorithms and compared their performance to a baseline linear feedback algorithm. Using our benchmarking testbed, we showed that the DMPC algorithms outperform the linear feedback algorithm for a platoon of four vehicles. We performed further experiments in simulation to demonstrate that not only do the DMPC algorithms perform better on hardware, they also scale well to increases in size of the platoon.

Future work will consider how we can improve upon these methods both in theory and practice. Theoretically, there is a need to ensure the DMPC algorithms we use are string stable, and not just asymptotically stable. This may not be possible with the current formulation, but a similar formulation with a CTH policy has potential. In this work, we arbitrarily select desired distances, but future research could investigate how these distances can be rigorously determined to ensure safety (\ie, no collisions) and performance (\ie, small inter-vehicle spacing). Finally, future research may explore the benefits of incorporating provably safe machine learning methods and benchmarking against the control-theoretic methods used in this paper.

\section*{Acknowledgements}

Research was sponsored by the DEVCOM Analysis Center and was accomplished under Cooperative Agreement Number W911NF-22-2-001. The views and conclusions contained in this document are those of the authors and should not be interpreted as representing the official policies, either expressed or implied, of the Army Research Office or the U.S. Government. The U.S. Government is authorized to reproduce and distribute reprints for Government purposes notwithstanding any copyright notation herein.

\bibliographystyle{abbrv}
\bibliography{refs}

\end{document}